\def\year{2022}\relax
\documentclass[letterpaper]{article} 
\usepackage{aaai22}  
\usepackage{times}  
\usepackage{helvet}  
\usepackage{courier}  
\usepackage[hyphens]{url}  
\usepackage{graphicx} 
\usepackage{natbib}  
\usepackage{caption} 
\DeclareCaptionStyle{ruled}{labelfont=normalfont,labelsep=colon,strut=off} 
\usepackage{algorithm}
\usepackage{algorithmic}

\usepackage{newfloat}
\usepackage{listings}
\floatstyle{ruled}
\newfloat{listing}{tb}{lst}{}
\floatname{listing}{Listing}
\newcommand{\approach}{{\sc RetroNLU}}

\usepackage{times}
\usepackage{latexsym}
\usepackage{float}

\usepackage[utf8]{inputenc}

\usepackage[T1]{fontenc}
\usepackage{microtype}
\usepackage{paralist}

\usepackage{enumitem}
\usepackage{xspace}

\usepackage{booktabs} 
\usepackage{wasysym} 
\usepackage{array}

\usepackage{tikz}
\usepackage{pgfplots}
\usepackage{float}
\usepackage{arydshln}
\usepackage{centernot}
\usepackage{bbm}

\newcommand{\exampleParagraph}[2]{\framebox{\parbox{0.95\linewidth}{\paragraph{#1} 
{\footnotesize #2}}}}

\title{\approach: Retrieval Augmented Task-Oriented Semantic Parsing}
\author {
    Vivek Gupta\textsuperscript{\rm 1,2\thanks{Work done by author while interning at Facebook Conversational AI.}},
    Akshat Shrivastava\textsuperscript{\rm 2},
    Adithya Sagar\textsuperscript{\rm 2},
    Armen Aghajanyan\textsuperscript{\rm 2},
    Denis Savenkov\textsuperscript{\rm 2}
}
\affiliations {
    \textsuperscript{\rm 1}School of Computing, University of Utah \\\textsuperscript{\rm 2}Facebook Conversational AI, Menlo Park\\
    vgupta@cs.utah.edu 
    \{akshats, adithyasagar, armenag, denxx\}@fb.com 
}
\usepackage{bibentry}
\usepackage{pgfplots}
\pgfplotsset{compat=1.7}

\begin{document}

\maketitle

\begin{abstract}
While large pre-trained language models accumulate a lot of knowledge in their parameters, it has been demonstrated that augmenting it with non-parametric retrieval-based memory has a number of benefits from accuracy improvements to data efficiency for knowledge focused tasks, such as question answering.
In this paper we are applying retrieval-based modeling ideas to the problem of multi-domain task-oriented semantic parsing for conversational assistants.
Our approach, \approach, extends a sequence-to-sequence model architecture with retrieval component, used to fetch existing similar examples and provide them as an additional input to the model.
In particular, we analyze two settings, where we augment an input with (a) retrieved nearest neighbor utterances (utterance-nn), and (b) ground-truth semantic parses of nearest neighbor utterances (semparse-nn).
Our technique outperforms the baseline method by 1.5$\%$ absolute macro-F1, especially at the low resource setting, matching the baseline model accuracy with only 40\% of the data.
Furthermore, we analyze the nearest neighbor retrieval component's quality, model sensitivity and break down the performance for semantic parses of different utterance complexity.
\end{abstract}

\section{Introduction}
\label{sec:introduction}

\begin{figure}[!h]
    \centering
    \includegraphics[width=0.35\textwidth]{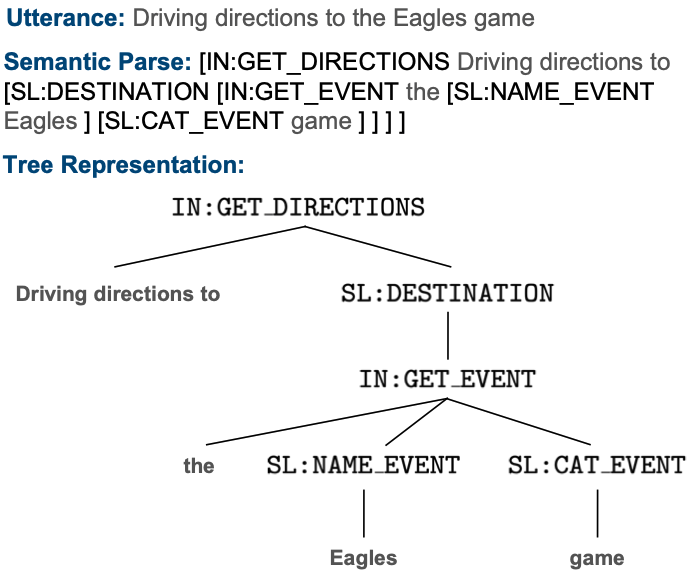}
    \caption{An intent-slot based compositional semantic parsing example (coupled) from TOPv2 \cite{chen2020low}.}
    \label{fig:semparse_example}
\end{figure}

\citet{roberts2020knowledge} demonstrated that neural language models quite effectively store factual knowledge in their parameters without any external information source.
However, such implicit knowledge is hard to update, i.e. remove certain information \cite{bourtoule2019machine}, change or add new data and labels.
Additionally, parametric knowledge may perform worse for less frequent facts, which don't appear often in the training set, and ``hallucinate'' responses.
On the other hand, memory-augmented models \cite{sukhnaatar2015mem} decouple knowledge source and task-specific ``business logic'', which allows updating memory index directly without model retraining.
Recent research demonstrated a potential of such approaches for knowledge-intensive NLP tasks, such as question answering \cite{Khandelwal2020Generalization,lewis2020retrievalaugmented}.
In this work, we explore \approach: retrieval-based modeling approach for task-oriented semantic parsing problem, where explicit memory provides examples of semantic parses, which model needs to learn to transfer to a given input utterance.
An example semantic parse for task-oriented dialog utterance and its corresponding hierarchical representation are presented in Figure \ref{fig:semparse_example}. 

In this paper we are focusing on the following questions:
\textit{Data Efficiency: can retrieval based external non-parametric knowledge reduce dependency on parametric knowledge usually learned from supervised training on large labeled dataset?}. \footnote{Parametric knowledge refers to the knowledge held in model parameters. Non-parametric knowledge, on the other hand, refers to external information sources that the model utilizes for inference.}
We analyse how several training settings based on amount of supervision data, i.e. fully supervised training vs limited supervised training, affect model prediction. 
\textit{Semi-supervised setting. Can we use abundant and inexpensive unlabeled external non-parametric knowledge instead of structurally labeled one for model improvement?}
We analyse the effect of using unlabeled similar utterances instead of labelled semantic parses as external non-parametric knowledge on model performance. 
\textit{Robustness to noise. Can a model robustly choose to use parametric knowledge instead of non-parametric knowledge for example with unreliable non-parametric knowledge?} 
We analyse the robustness of the model and it's dependence on the external non-parametric knowledge. One doesn't always have accurately labeled reliable external knowledge for all examples/utterances. 
\textit{Is the external non-parametric knowledge  addition effective for both rare and complex structured (hierarchical) examples too?}. 
We also analyse where the external knowledge addition is more fruitful, or does it support correct predictions for all examples equally. It would be interesting to see if external knowledge could also improve challenging and complex examples/utterances. 
Finally, we analyse the limit up-to which such external knowledge could be useful. We analyse the issues of structural redundancy in the nearest neighbors. \textit{Is addition of more and more external knowledge useful, or there are some limitations and challenges?.}

Our contribution in the paper are as follows:

\begin{enumerate}
    \item We show how combination of parametric and non-parametric knowledge can enhance the complex structured NLP task of task oriented semantic parsing.
    \item We show the effectiveness of this approach in important scenario of limited labeled data training (i.e. limited parametric knowledge).
    \item We demonstrate a potential of retrieval-based modeling in semi-supervised settings, where the input to the model is augmented with unannotated examples.
    \item We show the model robustness to external knowledge quality (retrieval) by comparing model prediction on clean vs noisy external non-parametric knowledge.
    \item Finally, we analyze performance improvements on inputs of various complexity: output semantic structure complexity and input frequency (i.e. frequent/rare).
\end{enumerate}

The dataset, as well as the accompanying scripts, will be made available soon.

\section{Proposed Approach}
\label{sec:problem&solution}

\begin{figure*}[ht!]
    \centering
    \includegraphics[width=0.73\textwidth]{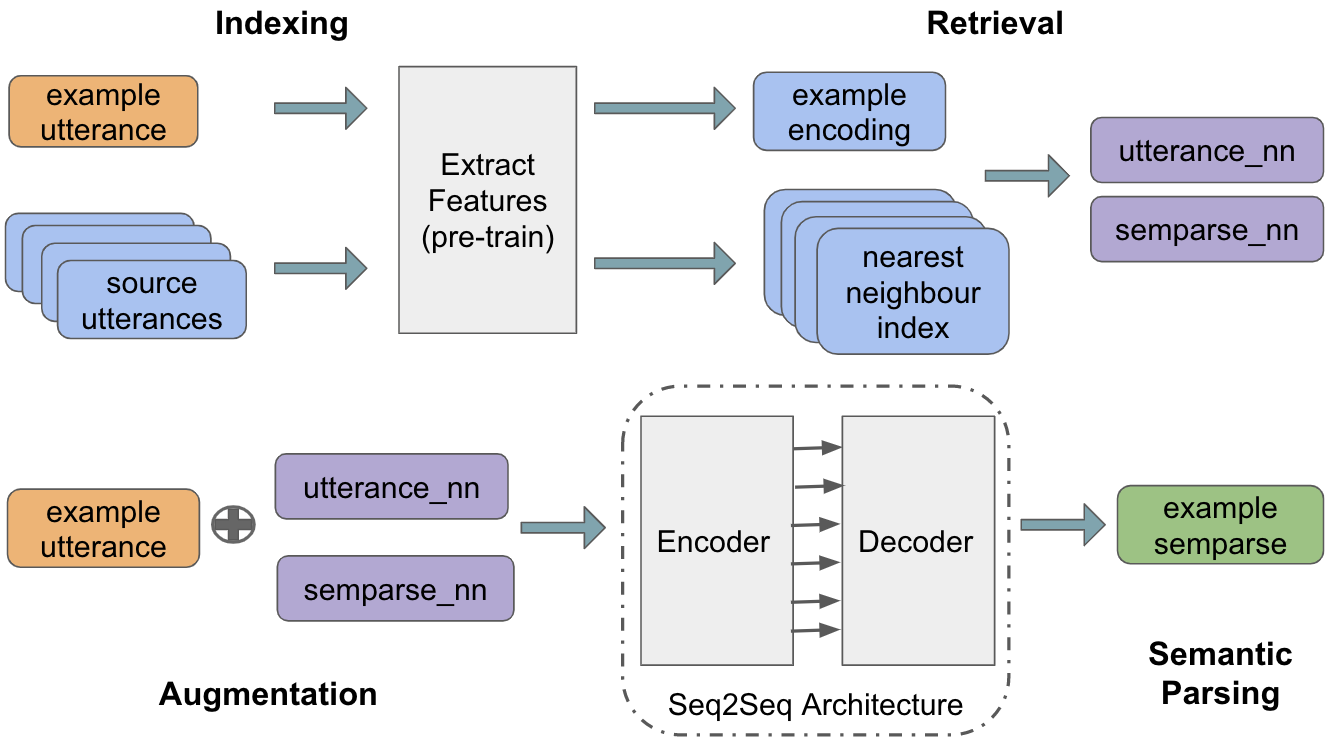}
    \caption{High level flowchart for retrieval augmented semantic parsing (\approach) approach.}
    \label{fig:flowchart}
\end{figure*}

Our proposed approach has four main steps \begin{inparaenum}[(1)] \item \textbf{building index} by embedding text of the training examples; \item \textbf{retrieval} given an example utterance, retrieve nearest neighbor utterances from the index. \item \textbf{augmentation: } append the nearest-neighbor utterance ground truth semantic parse (semparse-nn) or utterance itself (utterance-nn) to the original input using a special separator token (such as `\texttt{|}'), and \item \textbf{semantic parsing}: use the retrieval-augmented input and ground truth as output to train a semantic parsing model.\end{inparaenum}

Figure \ref{fig:flowchart} shows high level flowchart for the Retrieval Augmented Semantic Semantic Parsing (\approach) approach.

\paragraph{Indexing: } To build an index we use a pre-trained BART model to get training utterance embeddings. 
More specifically, we get sentence embedding for all the training utterances. These sentence embeddings are obtained as average of token embeddings from last model layers. \footnote{ extract\_features function \url{https://bit.ly/3trOuRj}} We then used this embedding to build a fast and efficient retrieval index \cite{JDH17}. 
\paragraph{Retrieval: } Next, given a new input (training or test row), we obtain embeddings by running it through same pre-trained BART, and then query the index with it to retrieve nearest neighbors text and their ground truth semantic parses. 
For training data, we exclude an example itself from the retrieved list.
For example, let the input utterance be \textit{`please add 20 minutes on the lasagna timer'}, and then the nearest neighbour \textit{`add ten minutes to the oven timer'} with the semantic parse as \textit{`[in:add$\_$ time$\_$ timer add [sl:date$\_$ time ten minutes ] to the [sl:timer$\_$ name oven]  [sl:method$\_$ timer timer] ]'}. 

\paragraph{Augmentation: }
Once we got a list of nearest neighbors, we can append either utterance text or semantic parse to the input, following the left to right order.
The closest neighbor appears to the immediate left of the input example utterance.
One can also directly append the nearest neighbor utterance rather than the semparse, refer as utterance-nn. For the last example the final input would after augmentation would be \textit{'[in:add$\_$ time$\_$ timer add [sl:date$\_$ time ten minutes ] to the [sl:timer$\_$ name oven]  [sl:method$\_$ timer timer ] ] $|$ please add 20 minutes on the lasagna timer'} for semparse-nn, and \textit{'add ten minutes to the oven timer $|$ please add 20 minutes on the lasagna timer'} for utterance-nn. Here, the token `|' act as a separator between the input utterance and the neighbour's. 

\paragraph{Semantic Parsing: } 
The final step is to train a sequence-to-sequence model (e.g. LSTM or Transformer). We fine-tune a BART model with copy mechanism \cite{aghajanyan-etal-2020-conversational}, which incorporates benefits of pre-trained language model (BART) and sequence copy mechanism (copy-ptr), and most importantly obtain state-of-the-art results on TOPv2 \cite{chen2020low}. The retrieval augmented example is an input to the encoder and the corresponding ground-truth semantic parse as the labeled decoded sequence. 

At test time, we simply pass the augmented input to the trained \approach model, and take it's output as the predicted semantic parse for the given utterance as input.

\section{Experiment and Analysis}
\label{sec:experiment&analysis}

Our experiments are designed to investigate the effect of our knowledge-retrieval-based augmentation strategy on model performance measures such as accuracy and data efficiency. Our experiments studies the following questions :

\begin{enumerate}
    \item Can today's large pre-trained models exploit the non-parametric retrieval information as described in \S\ref{sec:problem&solution} to enhance task-oriented semantic parsing?
    
    \item Does non-parametric memory benefit examples equally across the distribution, e.g. do we observe larger improvements for more complex and less frequent semantic frames compared to simpler head examples.
    
    \item Does augmentation with retrieved unannotated examples (utterance\_nn) improve semantic parsing accuracy, which represents a semi-supervised setting when only part of the dataset is provided with semantic parses?
    
    \item How sensitive is the model to noise in the retrieval (nearest neighbor) information? Can the model predict correct intent/slots for examples with poor retrieval quality?

    \item What is the data efficiency of a retrieval-augmented semantic parsing model? Does it achieve higher accuracy with less training data compared to the baseline sequence-to-sequence model?
    
\end{enumerate}

Overall, our experiments' aims is to demonstrate that non-parametric external knowledge can benefit a parametric model \footnote{We did not seek to change the existing architecture because we want the augmentation technique to be adaptable to other methods.}, and to conduct an in-depth analysis of the effects.

\subsection{Experimental setup} 

In this section, we discuss the datasets, pre-processing, and the model used throughout the experiments.

\paragraph{Datasets} For our experiments we used the Task-Oriented Semantic Parsing TOPv2 \cite{chen2020low} dataset for dialog systems. We concentrated our efforts on task-oriented parsing because of the commercial importance of data efficiency requirements. \footnote{Despite the fact that the structure of the augmented neighbors may change, the augmentation strategy should remain the same.}
The TOPv2 dataset contains utterances and their semantic representations for 8 domains: source domains - alarm, messaging, music, navigation, timer, and event, and target domains: reminder and weather, designed to test zero-shot setting. 
For our experiments we chose source domains, which has a good mixture of simple (flat) and complex (compositional) semantic frames. For dataset statistics refer Table 1 in \cite{chen2020low}. 

\paragraph{Data Processing}
To build a retrieval index we used the training split of the dataset.
Each utterance was represented by its BART-based embedding and indexed using FAISS library \cite{johnson2019billion}. \footnote{We use L2 over unit norm BART embedding for indexing.}
To produce augmented examples, we retrieved nearest neighbors for each training and test examples, excluding exact matches.
In the augmented examples, we use the special token `\texttt{$|$}' to separate the nearest neighbors, as well as utterance with the first neighbor.\footnote{Using different separator tokens for neighbor-neighbor pair and utterance-neighbor pair didn't improve performance.} We used only one neighbor for most experiments except when we analyse the effect of multiple neighbors on the model performance. 

In nearest neighbor augmented input, we followed right to left order, where the actual model input comes last, and its highest ranked neighbor is appended to the left of the utterance, followed by other neighbors in the left based on ranking.\footnote{Left to right ordering with utterance in leftmost, followed the neighbors in index order has similar performance.} For input data preprocessing, we follow \cite{chen2020low} procedure, we obtain BPE tokens \cite{papineni2002bleu} of all tokens, except ontology tokens (intents and slot labels), which are treated as atomic tokens and appended to the BPE vocabulary. Furthermore, we use the decoupled canonical form of sem-parse for all our experiments. For decoupling, phrases irrelevant to slot values are removed from sem-parse, and for canonicality, slots are arranged in alphabetic order \cite{aghajanyan2020conversational}.

\paragraph{Models}For fair comparison with the earlier baseline, we use the state-of-the-art BART based Seq2Seq-CopyPtr model for task-oriented semantic parsing. \footnote{We prefer transformer-based language model over  non-transformer models, such as LSTM, because the later does not capture extended context as well as the former.} The BART based Seq2Seq-CopyPtr model initialize both the encoder and decoder with pre-trained BART \cite{lewis2020bart} model. We choose the BART based Seq2Seq-CopyPtr model for the task because it's a strong baseline, the performance of the other language model such as RoBERTa without augmentation was inferior \cite{chen-etal-2020-low, aghajanyan-etal-2020-conversational}. \footnote{However, we believe that our findings also apply to other models, including RoBERTa, because our conclusions are general.} The model is using the copy mechanism \cite{see2017get}, which enables it to directly copy tokens from the input utterance (or from example semantic parses from nearest neighbors). For training we use $100$ epochs, Adam optimizer \cite{kingma2014adam} with learning rate $\alpha$ of $1e-4$ and decay rate $\beta_1$ and $\beta_2$ of $0.9$ and $0.98$ respectively in all our experiments. Also, we didn't added any left or right padding and rely on variable length encoding in our experiments. We use warm-up steps of $4000$, dropout ratio of $0.4$, and weight decay $0.0001$, but no clip normalization as regularization during the training. We use batch size of $128$ and maximum token size of $2048$. Furthermore, to ensure both encoder and decoder BART, can utilise the extra nearest neighbour information, we increase the embedding dimension to $1024$. We use the same hyper-parameters for all model training, i.e. baseline (without-nn) and \approach~ models  (utternce-nn and semparse-nn).


\subsection{Results and Analysis}


The following sections describe the results of our experiments with respect to the research questions outlined above.

\paragraph{Full Training Setting} To verify if today's large pre-trained models exploit the non-parametric retrieval information as described in \S\ref{sec:problem&solution} to enhance task-oriented semantic parsing, we compare performance of original baseline (without-nn) with our retrieval augmented models, i.e. augmenting first neighbour utterance (utterance-nn) and augmenting first neighbour semantic parse (semparse-nn).
Table \ref{tab:mainresult} compares the frame accuracy of retrieval augmented \begin{inparaenum}[(a)] \item top nearest neighbour utterance (utterance-nn), \item top nearest neighbour ground-truth semantic parse (semparse-nn) \end{inparaenum} with original baseline (without-nn) with model train on complete training data. 

\textit{Analysis}: We observe performance improvements with retrieval-augmented models for most domains compared to the original baseline in both cases. The increase in performance (micro-avg) is more substantial 1.30 $\%$ with semparse-nn compare to 0.85 $\%$ with utterance-nn. The improvement in utterance-nn augmentation performance is likely due to memorization-based generalization, as explained earlier by \cite{khandelwal2019generalization}. \footnote{The scores are averaged over three runs with std. of $~0.3\%$} The results shows the retrieval augmented semantic parsing is overall effective. Furthermore, the performance enhancement can be obtained also with unstructured utterance (utternace-nn) as nearest neighbour. The utterance-nn based augmentation is particularly beneficial in semi-supervised scenarios, where we have a larger unlabelled dataset.

\begin{table}
\setlength{\tabcolsep}{3.0pt} 
\renewcommand{\arraystretch}{1} 
\small
    \centering
    \begin{tabular}{c|c|c|c}
    \hline
    \bf Domains &\bf  without-nn &\bf  utterance-nn & \bf semparse-nn \\ \hline
    Alarm & 86.67 & 87.17 &\bf 88.57 \\ 
    Event & 83.83 &\bf 85.03 & 84.77 \\  
    Music & 79.80 &\bf 80.73 &\bf 80.71 \\ 
    Timer & 81.21 &\bf 81.75 & 81.01 \\
    Messaging & 93.50 & 94.52 &\bf 94.65 \\
    Navigation & 82.96 & 84.16 &\bf 85.20 \\  \hline
    micro-avg & 84.43 & 85.28 &\bf 85.74 \\ 
    macro-avg & 84.66 & 85.56 &\bf 85.82 \\ \hline
    \end{tabular}
    \caption{Performance of \approach~ w.r.t original baseline (without-nn) with full training.}
    \label{tab:mainresult}
\end{table}

\begin{figure}[!htbp]
    \centering
    \includegraphics[width=0.45\textwidth]{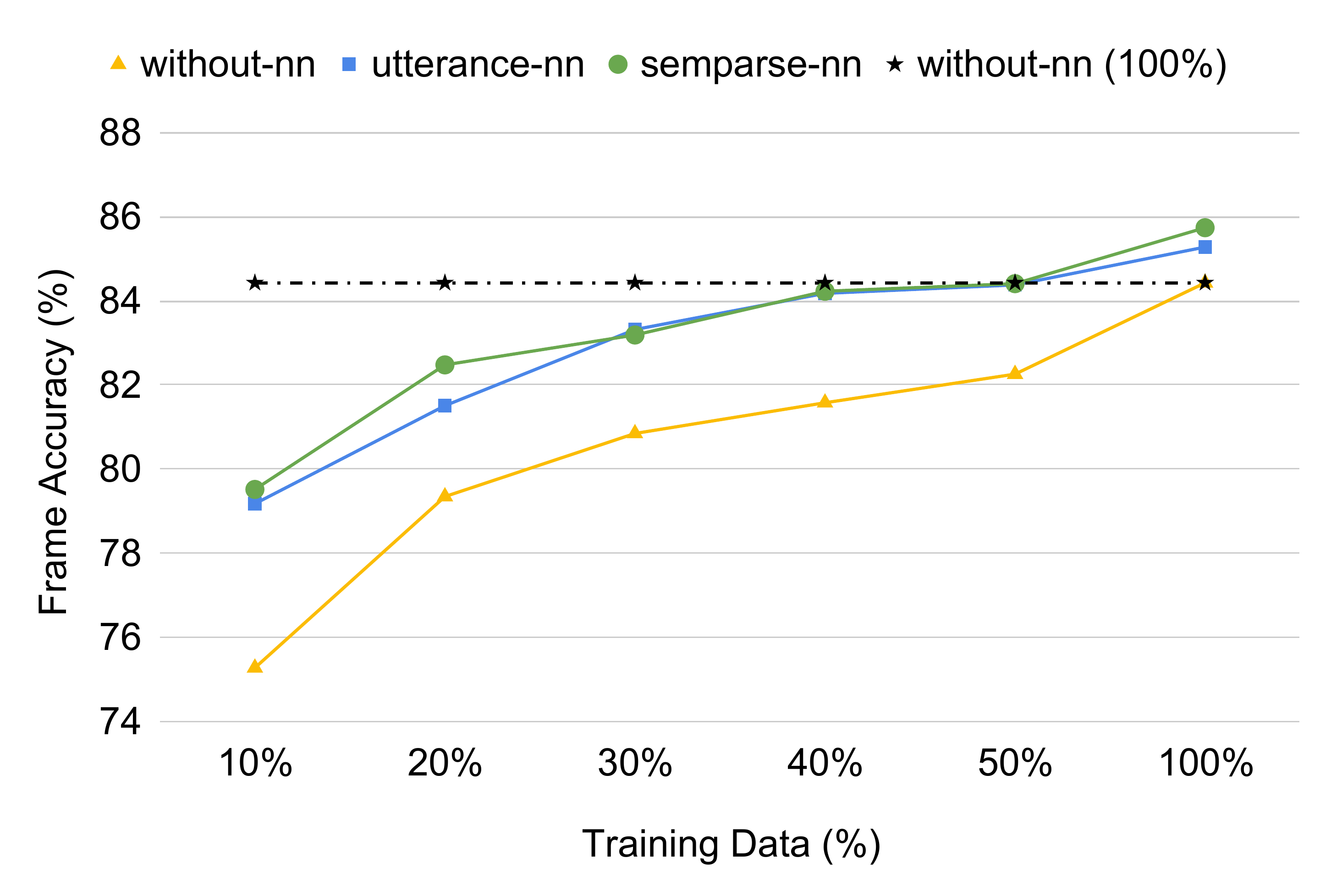}
    \caption{Performance of \approach~ w.r.t orignal baseline with limited supervised training.}
    \label{fig:limited_train}
\end{figure}

\paragraph{Limited Training Setting} To verify if \approach~ approach has better data efficiency, we compare performance of models which are trained with limited training data.
Figure \ref{fig:limited_train} shows frame accuracy (micro-avg) when we use only 10 $\%$ to 50$\%$ of the training data. The training datasets are created in an incremental setting so that next set include examples from the former set.

\textit{Analysis}: As expected, the performance of all models increases with training set size. Both retrieval augmented models i.e. utterance-nn and semparse-nn outperform the without-nn baseline for all the training sizes.
The improvement via augmentation is more substantial with less training data, i.e. 4.24$\%$ at 10$\%$ data vs 1.30$\%$ at 100$\%$ data.
Furthermore, the semparse-nn augmented model outperforms the original-full train without-nn model with only 40$\%$ of the data. 
The results show that the retrieval augmented semantic parsing is more data efficient, i.e. when there is  \begin{inparaenum}[(a)] \item limited labelled training dataset with more unlabelled data for indexing (utterance-nn), and \item sufficient training data but limited training time (seqlogical-nn).\end{inparaenum}

The first case is useful when the ground truth label is missing for utterances due to lack of annotation resources. 
In such a scenario, one can build the index using large amount of unlabeled utterances and use the index for augmentation. The second case helps us train the model faster, while maintain all annotated examples in the index.
In such a case, one can update the retrieval index only, without re-training the model again and again.
This is useful when training on full data is not possible due to limited access to model (black-box), a cap on the computation resources available, or for saving training time i.e. industries fast deployment need.

\begin{figure}[!htbp]
    \centering
    \includegraphics[width=0.45\textwidth]{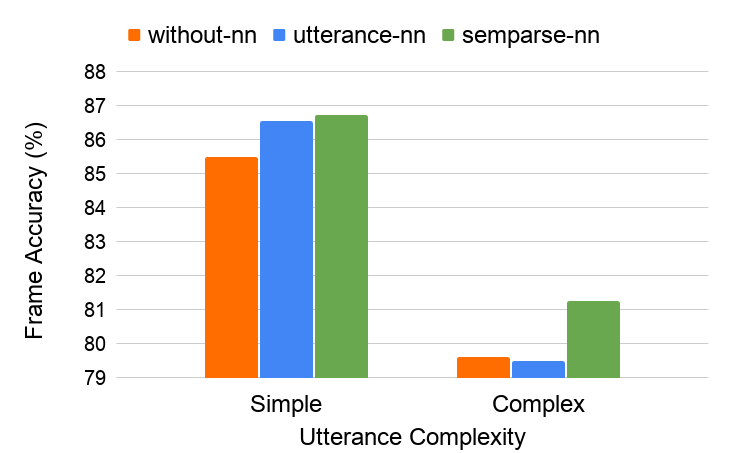}
    \caption{Performance comparison (micro-avg) of \approach~ w.r.t original baseline (without-nn) with utterance complexity, i.e. simple and complex.}
    \label{fig:utteranceComplex}
\end{figure}

\paragraph{Effect of Utterance Complexity} To verify if retrieval-based model work well for hard examples in addition to easy examples, we analyse the retrieval augmented model performance improvements (with full training) on simple utterance with only one level in semantic representation (depth-1) vs complex utterance with hierarchical semantic frames (compositional depth-2 and above). Figure \ref{fig:utteranceComplex} shows frame accuracy of without-nn, utterance-nn and semparse-nn model with utterance complexity.

\begin{figure}[!htbp]
    \centering
    \includegraphics[width=0.45\textwidth]{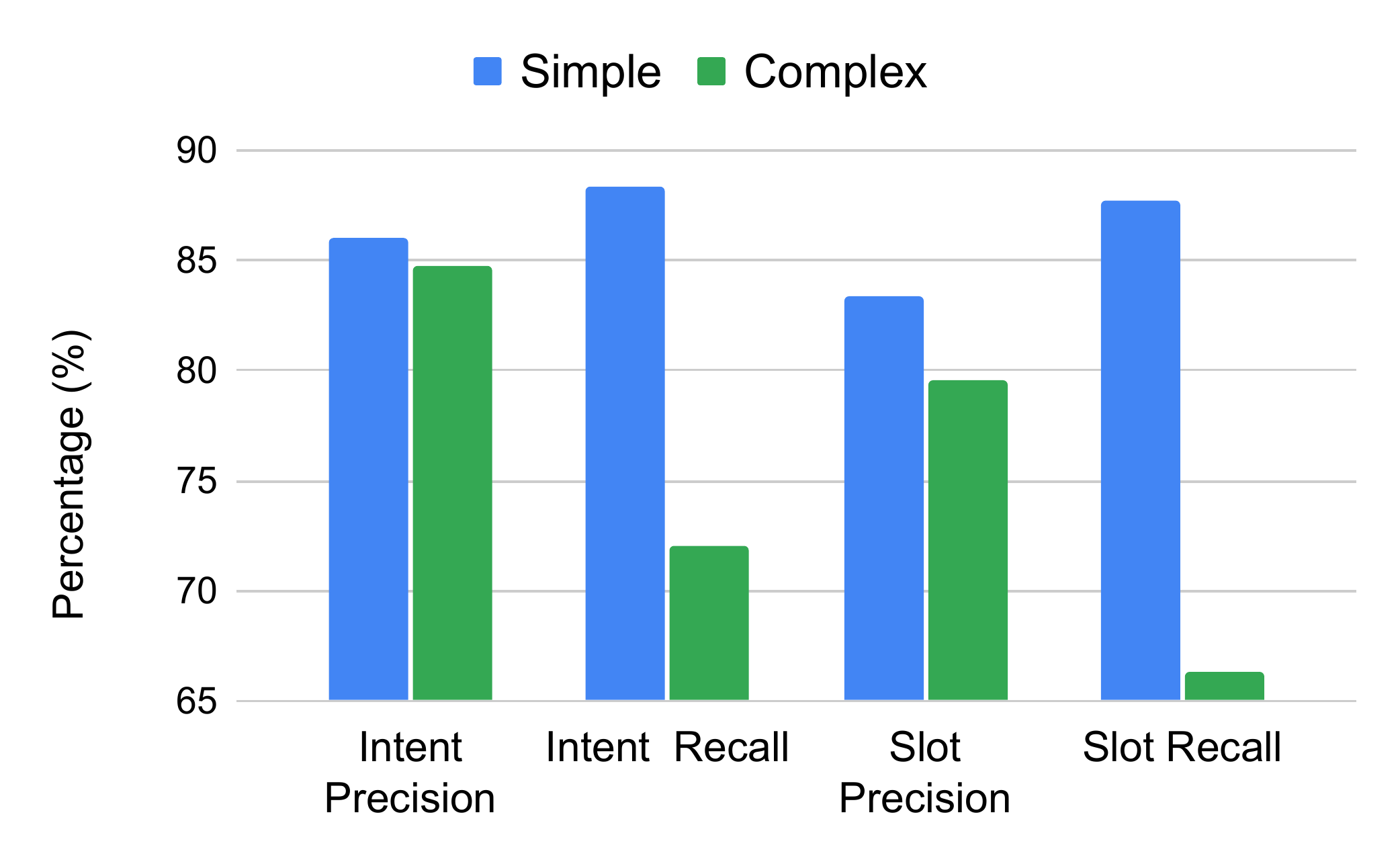}
    \caption{Precision and Recall of intents and slots for semparse-nn nearest neighbour w.r.t to gold semparse for the \approach.}
    \label{fig:utteranceComplexPrecRec}
\end{figure}

\textit{Analysis}: As expected, all models perform relatively poorly on complex utterances (79.5$\%$) in comparison to simple utterances (85.5$\%$). 
Interestingly, both augmentation models equally improve performance on simple queries.
And with semantic-frame based augmentation we observe a substantial performance improvement on challenging complex utterances, of 2$\%$, with respect to the original baseline (without-nn). 
This suggests, that by retrieving nearest neighbors and providing a model with examples of complex parses, the model learns to apply it to a new request. Figure \ref{fig:utteranceComplexPrecRec} shows precision and recall for intents and slots in retrieved semantic parses.
The recall for intent and slot retrieval is 15$\%$ lower for complex utterances. \footnote{The gap for precision was not substantial 1$\%$ for intents and  4$\%$ for slots.} Thus, highlighting one reason for a performance gap between simple and complex frames.

\begin{figure}[!htbp]
    \centering
    \includegraphics[width=0.45\textwidth]{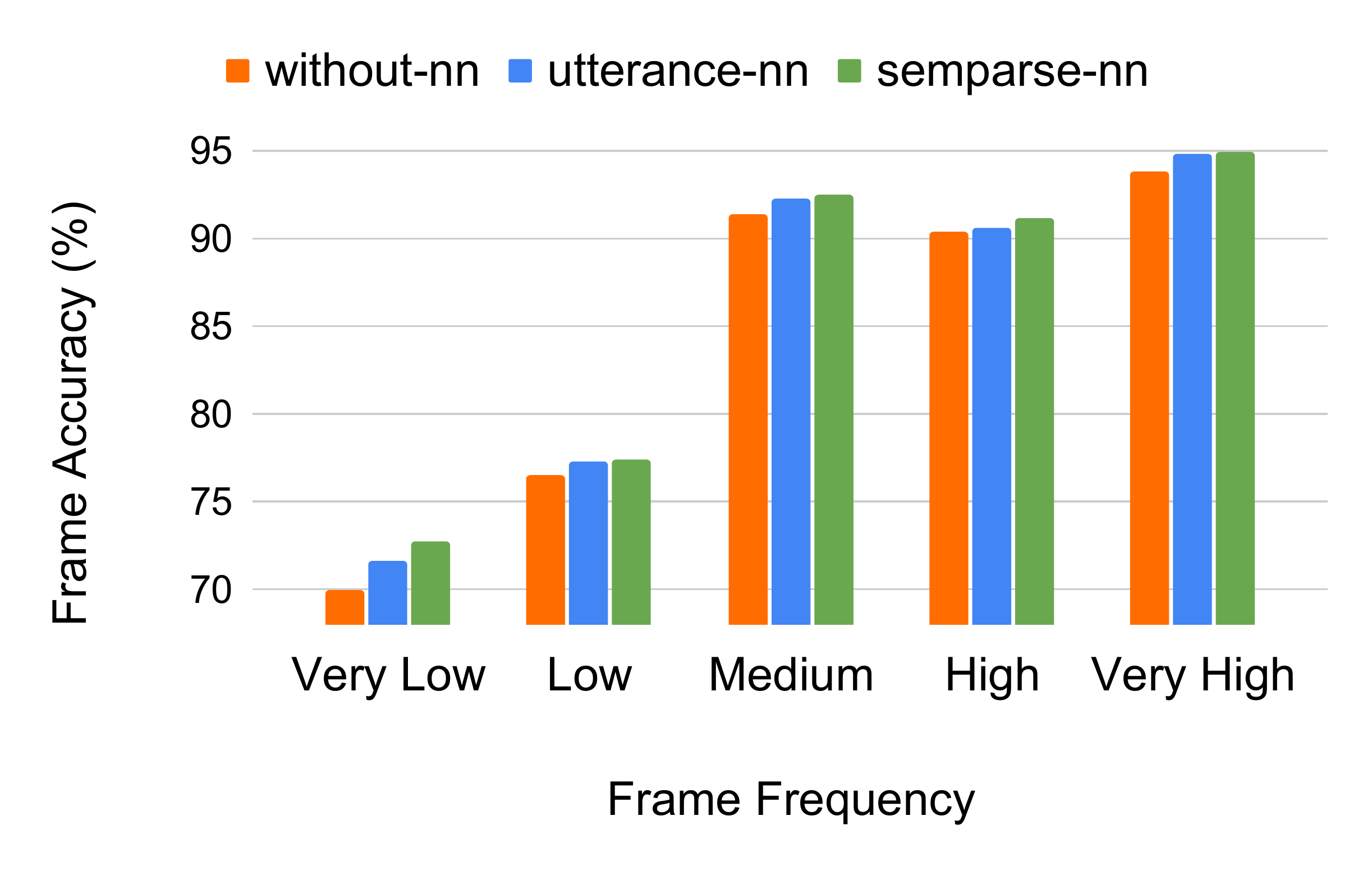}
    \caption{Performance of \approach~ w.r.t original baseline (without-nn) with varying frame frequency.}
    \label{fig:frequency}
\end{figure}

\paragraph{Effect of Frame Rareness} To verify if retrieval-based model is better at the tail, i.e. rare utterances, we analyze the retrieval augmented model performance improvement (with full training) with frame rareness, as shown in Figure \ref{fig:frequency}. 
Rare frames are those example utterances whose ground truth semantic parses without slot value tokens appear infrequently in the training set.
To analyze this, we divided the test set into five equal sizes i.e., Very Low, Low, Medium, High, and Very High sets, based on the frequency of semantic frame structure.
The experiment checks if performance improvement is mainly attributed to frequently repeating frames (frequent frames) or for rarely occurring frames (rare frames). Figure~\ref{fig:frequency} shows the main results from the experiment.

\begin{figure}
    \centering
    \includegraphics[width=0.45\textwidth]{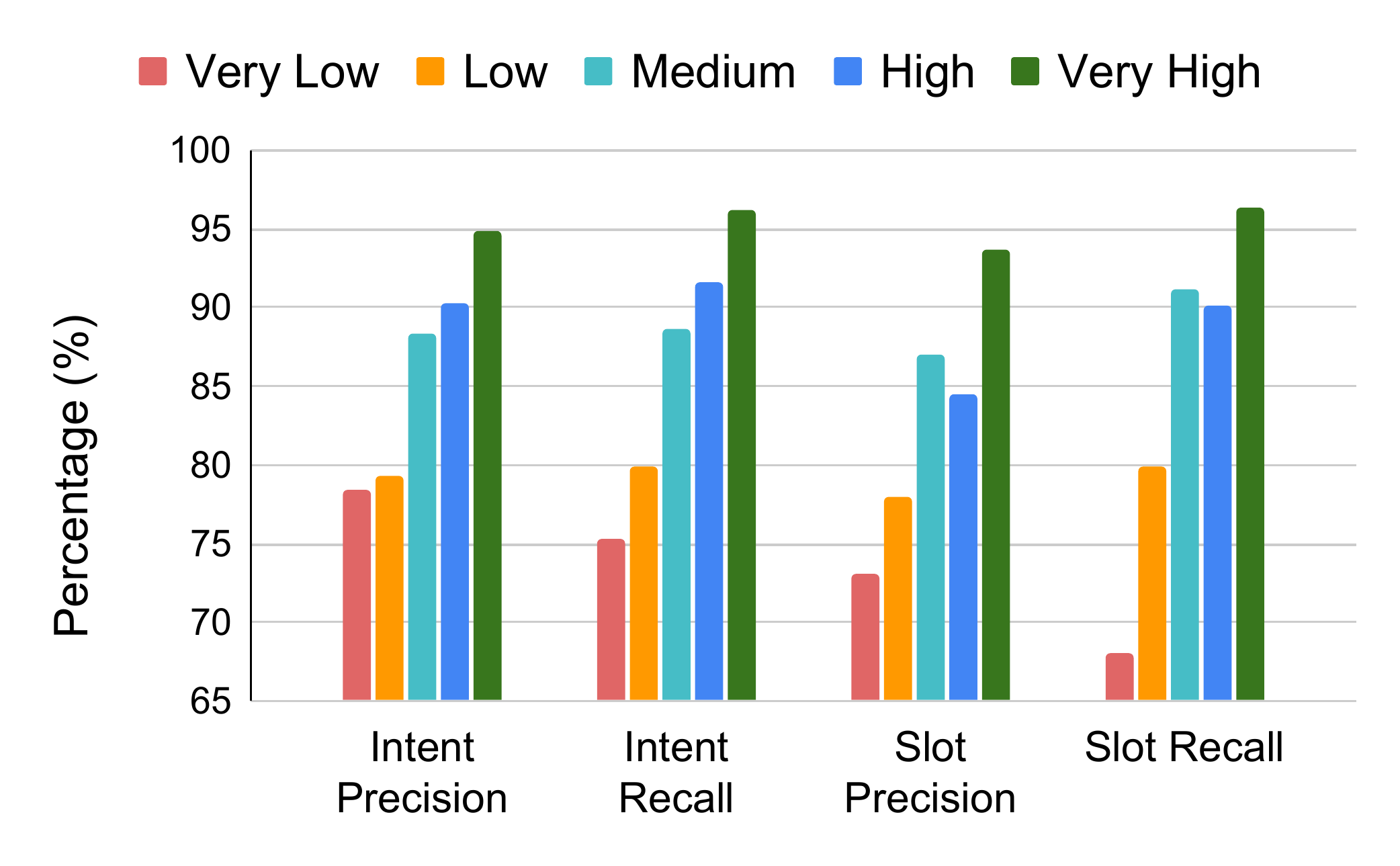}
    \caption{Precision and Recall of intents and slots w.r.t to frame frequency for semparse-nn of the \approach.}
    \label{fig:PrecRecallfrequency}
\end{figure}

\textit{Analysis}: Figure~\ref{fig:frequency} shows that all models perform worse on rare frames.
This is expected as the parametric model gets less data for training on these frames. Furthermore, many of the low-frequency frames are also complex utterances with more than one intent and have more slots too.
Moreover, the nearest neighbour will be noisier for less frequent frames.
This is evident from the lower values of precision (20$\%$ gap) and recall (25$\%$ gap) on the intent and slots for nearest neighbor as shown in the Figure \ref{fig:PrecRecallfrequency}. 

\begin{figure}
    \centering
    \includegraphics[width=0.45\textwidth]{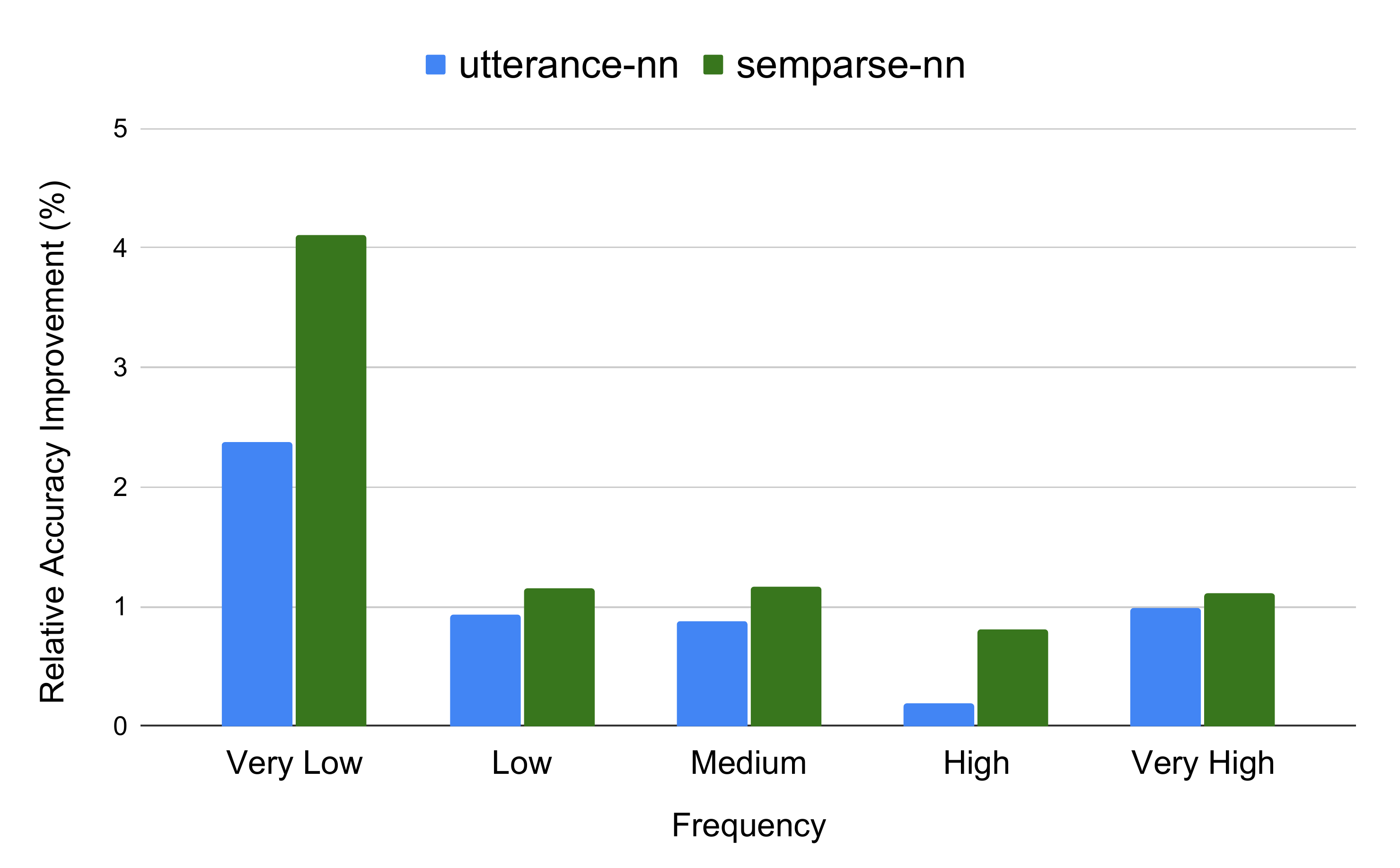}
    \caption{Relative performance improvement of \approach~ w.r.t original baseline (without-nn) with varying frame frequency.}
    \label{fig:relative_improvement}
\end{figure}

However, compared to original baseline (without-nn) the relative performance improvement on rare frames with retrieval augmented model is more substantial, as shown in Figure \ref{fig:relative_improvement}.
For example, the relative improvement for Very Low frequency frames is 2.37$\%$ (utterance-nn) and  4.11$\%$ compared to just 1.01$\%$ (utterance-nn) and 1.11 $\%$ for the Very High Frequency frames.
We suspect this is because of the model's ability to copy the required intent and slots from nearest neighbors if the parametric knowledge fails to generate it.
This shows the retrieval augmented model is even more beneficial for the rare frames.
Similar to earlier observations, semparse-nn is better than utterance-nn for the augmentation.

\paragraph{Effect of the number of neighbors} Can we further improve the performance of our retrieval augmented model by adding more nearest neighbors?
We compare k = 1, 2, and 3 nearest neighbours for both utterance-nn and semparse-nn setups\footnote{Going beyond 3 neighbours was not beneficial, (a) due to 512 tokenization limit of BART, (b) exponential increase in training time, and (c) insignificant improvement in performance after 3 neighbours.}. The result of the experiment is reported in Table \ref{tab:numeighbor}.

\textit{Analysis}: As shown in Table \ref{tab:numeighbor} the model performance only improves marginally with more nearest neighbors.
We attribute this to the following two reasons \begin{inparaenum}[(a)] \item redundancy - many utterance examples can share the same frame, as evident from the high accuracy for frequent frame Figure \ref{fig:frequency}., and \item complexity - as $k$ increases, the problem is getting harder for the model with longer inputs, more irrelevant and noisier inputs. \end{inparaenum}
To further verify the above reasons, we examine the semparse-nn retrieve nearest neighbor quality by comparing the intent and slot both Precision and Recall score for closest (k=1) and farthest (k=3) neighbor w.r.t to the gold semparse.
From Table \ref{tabl:retrievalquality} it is evident that precision and recall for intents and slots decrease as we go down the ranked neighbors list. 
Adding more nearest neighbour would only be beneficial when added neighbour capture diverse and different semantic structure which is missing from earlier neighbor and essential for the correct semparse.

\begin{table}[!htbp]
\small
\centering
\begin{tabular}{c|c|c|c}
\hline
\bf $\#$neighbors	&\bf  k = 1	&\bf k = 2 &\bf  k = 3 \\ \hline
without-nn	& 84.43	& 84.43	& 84.43 \\ 
utterance-nn & 85.28 & 85.35 & 85.40 \\ 
semparse-nn	&\bf 85.74	&\bf 85.81	&\bf 85.80 \\ \hline
\end{tabular}
\caption{Performance with increasing nearest neighbors.}
\label{tab:numeighbor}
\end{table}

\begin{table}[!htbp]
\small
    \centering
    \begin{tabular}{c|cc|cc}
    \hline
\bf Metric  & \bf Average & \bf Precision &  \bf Average & \bf Recall  \\ \hline
\bf Intent 		& \bf Farthest &\bf  Closest &\bf  Farthest  &\bf  Closest \\
Train & 81.39 &\bf 84.84 & 81.81 &\bf 85.04 \\
Valid & 80.46 &\bf 87.59 & 81.10  &\bf 87.93 \\
Test & 79.09 &\bf 86.23 & 79.35  &\bf 86.22 \\ \hline
\bf Slot	& \bf  Farthest &\bf  Closest &\bf  Farthest  &\bf  Closest \\
Train & 75.02 &\bf 80.05 & 79.56 &\bf 83.19 \\ 
Valid & 73.40 &\bf 82.38 & 79.77 &\bf 85.81 \\ 
Test & 74.59 &\bf 83.21 & 79.51 &\bf 85.11 \\ \hline
    \end{tabular}
    \caption{Precision / Recall of Intent-slots for \approach~ semparse-nn with closest / farthest  neighbor.}
    \label{tabl:retrievalquality}
\end{table}

\begin{figure}[!htbp]
    \centering
    \includegraphics[width=0.45\textwidth]{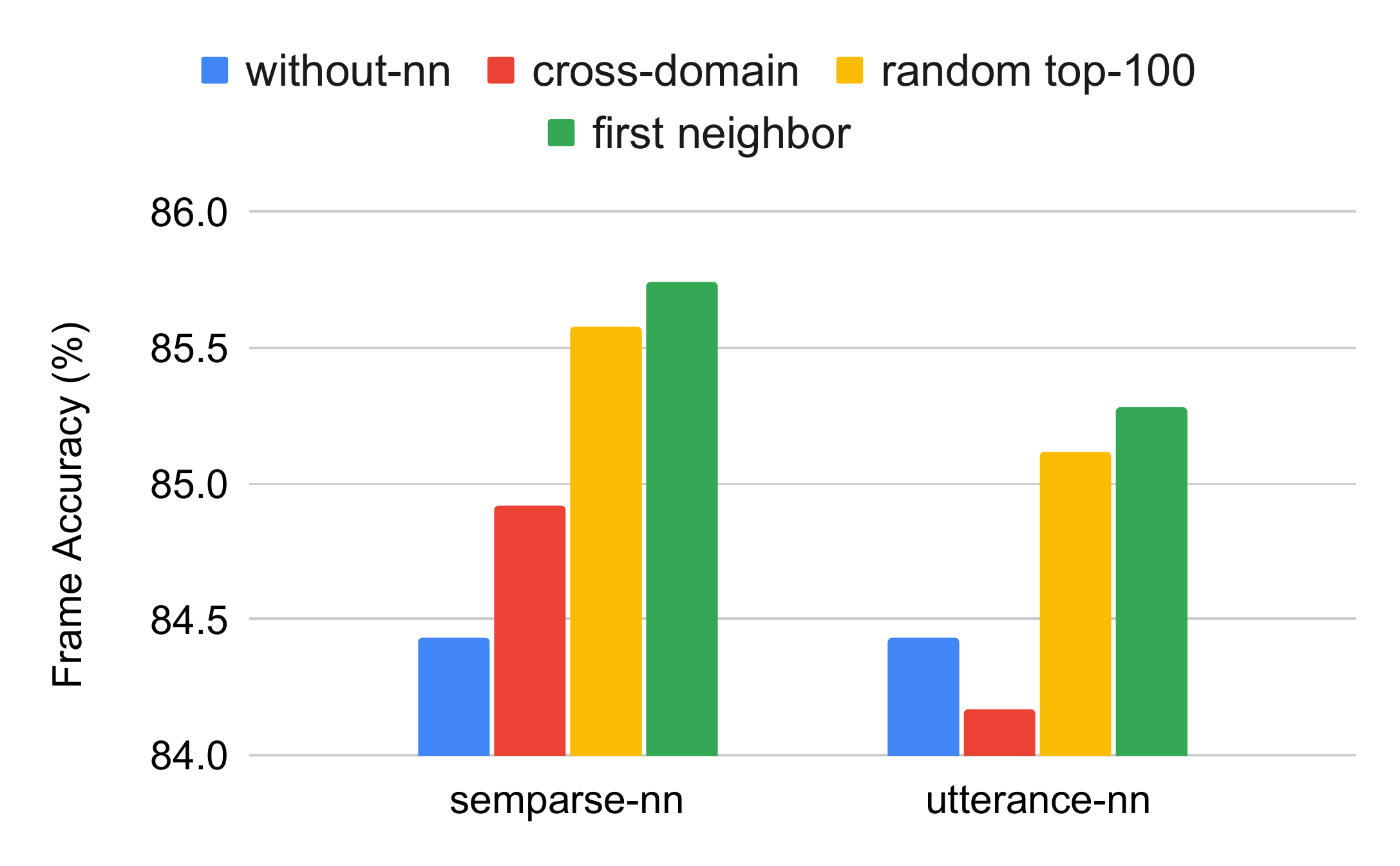}
    \caption{Performance of \approach~ with varying nearest neighbor quality on test data.}
    \label{fig:randneighbor}
\end{figure}

\paragraph{Effect of Retrieval Quality} To check if our \approach~ model is robust to the noise in the retrieved examples, we analyse the effect of quality of retrieval by comparing semantic parsing accuracy of top neighbor augmented models on the test data with \begin{inparaenum}[(a)] \item the top neighbour with random neighbor from domain other than the example domain, and \item random neighbor selected from top 100 ranked nearest neighbors in the index. It should be noted that these 100 top rank nearest neighbour can have some redundant semparse-nn structure, only slot values might differ. \end{inparaenum} Figure \ref{fig:randneighbor} shows the results of the experiments. 

\textit{Analysis}: From Figure~\ref{fig:randneighbor} it is clear that quality of nearest neighbor affect the semantic parsing accuracy.
We observe a 0.4$\%$ drop when random neighbor from top 100 nearest neighbor is chosen, instead of first neighbor, the small drop is because of redundancy in intent/slots structure between examples, only slots value could be major difference. However, the performance is still 0.9$\%$ to 1.0$\%$ better than the one without the nearest neighbor. We suspect this is because of the fact that the data has many utterances with similar semparse output. Upon deeper inspection we found that top-100 still includes many relevant frames, and therefore random examples from top-100 are often still relevant. Furthermore, there is also frame redundancy, many different utterance queries have similar semantic parse frames structure and only differ at the slot values. This is also evident from table \ref{tab:numeighbor} which shows adding more neighbors is not beneficial, because of frame redundancy. Surprisingly, we also observe that the model performance with random cross-domain neighbor is better than without-nn for semparse-nn by 0.5$\%$. This shows that the model knows when to ignore the nearest neighbor and when to rely on the parametric model. Furthermore, it also indicates that underlying parametric model parameters is improved by retrieval augmented training for the semparse-nn.

For the utterance-nn the performance drops when testing on cross-domain nearest neighbor augmented example. Thus, underlying the utterance-nn model is more sensitive than semparse-nn to the nearest neighbor quality. In addition, we also conducted an experiment in which we added the best possible neighbor based on the gold parse frame structure. The trained model, though this approach was not robust and relies too heavily on coping frames from neighbors, resulting in poor generalization. Our technique, on the other hand, with embedding-based retrieval, is good at generalization and has enhance the underlying parametric model. Overall, we can conclude that the semparse-nn and utterance-nn model are both quite robust to nearest neighbors quality. We can also conclude that the semparse-nn model was able to capture richer information through additional similar inputs than without-nn. However, to obtain the best performance good quality neighbour is an essential.

\section{Related Work}
\label{sec:related_work}

\paragraph{Task-oriented Semantic Parsing} 
Sequence-to-sequence (\emph{seq2seq}) models have recently achieved state of the art results in semantic parsing  \cite{rongali2020don, gupta2018semantic}, and they also provide a flexible framework for incorporating session-based, complex hierarchical semantic parsing \cite{sun2019knowledge,aghajanyan2020conversational,cheng2020conversational,mehri2020dialoglue} and multilingual semantic parsing \cite{li2020mtop,louvan2020recent}.
Architectures, such as T5 and BART \cite{raffel2019exploring,lewis-etal-2020-bart}, extended the recent breakthroughs with large pre-trained language models, and pushed the performance even further. Such models are quite capable of storing a lot of knowledge in their parameters \cite{roberts2020knowledge}, and in this work we explore the benefits of additional non-parametric knowledge in a form of nearest neighbor retrieval for the task of semantic parsing. By retrieving similar utterances, and providing them to the model as additional input, we allow the model to utilize these examples and learn to transfer semantic parses to new inputs, which has advantages for more rare inputs, as well as in data efficiency. To improve low resource seq2seq parsers \citet{chen-etal-2020-low} have proposed looking at meta learning methods such as reptile, and \citet{ghoshal2021learning} have introduced new fine-tuning objectives. Our approach is focused on non-architecture changes to augment generation with retrieval and thus can be combined with either of these approaches.

\paragraph{Incorporating External Knowledge}
An idea to help a model by providing an additional information, relevant to the task at hand is not new.
This includes both implicit memory tables \cite{weston2014memory,sukhbaatar2015end}, as well as incorporating this knowledge explicitly as an augmentation to the input. 
Explicit knowledge are incorporated in one of the following two ways \begin{inparaenum}[(1)] \item  suitable model architecture change to incorporate dedicated extended memory space internally i.e. memory network  \cite{bapna2019non,guu2020realm, lewis2020pre, tran2020cross}, and \item appending example specific extra knowledge externally with the input example directly without modifying model architecture \cite{papernot2018deep,weston2018retrieve, lewis2020retrieval, tran2020cross, khandelwal2020nearest, fan2021augmenting, chen2018neural,wang2019improving, neeraja-etal-2021-infotabskg} \end{inparaenum}. Retrieval-augmented approaches have been improving language model pre-training as well \cite{guu2020realm,lewis2020pre,tran2020cross}.
The idea here is to decouple memorizing factual knowledge and actual language modeling tasks, which can help mitigate hallucinations, and other common problems.

Multiple works like DkNN \cite{papernot2018deep}, RAG \cite{lewis2020retrieval}, kNN-LM \cite{tran2020cross}, kNN-MT \cite{ khandelwal2020nearest}, and KIF-Transformer \cite{fan2021augmenting} show that external knowledge is useful for large pre-trained language models, and can help fine-tuning. DkNN shows that nearest neighbour augmented transformer-based neutral network is more robust and interpretable. 
RAG shows that one can append external knowledge to improve open domian, cloze-style question answering, and even fact verification task such as FEVER.
kNN-LM shows that for cloze task for fact completion, one can combine nearest neighbour predictions with original prediction using appropriate weighting  to improve model performance. 
However, these works mostly study knowledge dependent question answering task, while we are exploring a different task of structural prediction of semantic frame structures for task-oriented dialog.

\section{Conclusion and Future Work}
\label{sec:conclusion}
In this work, we demonstrated that task oriented semantic parsing performance can be enhanced by augmenting parametric knowledge stored in neural models with non-parametric external memory.
We showed that appending examples retrieved from a nearest neighbor index help BART model with copy mechanism improve semantic parsing performance significantly on TOPv2 dataset.
Our \approach~ model is able to achieve higher accuracy earlier with less training data, which allows maintaining a large index with annotated data, while using only a subset to train a model more efficiently. Furthermore, we also observed improvements in semi-supervised settings, i.e. appending annotated similar retrieved examples also helps semantic parsing model achieve higher performance.
Lastly, we performed an analysis of performance improvements on different slices, and found \approach~ to be more effective on rarer complex frames, compared to a traditional \emph{seq2seq} model.

\approach~ extensions, we focus on joint training of retrieval and parsing components. Utterance representation should benefit from learning about the task at hand, and what utterances have similar semantic parse. We also wish to explore few/zero-shot performance of these models. Having an easy-to-update index allows you to modify annotations, add new or delete existing ones, with model potentially being sensitive to these changes. Finally, using cross-lingual representations like mBART \cite{liu-etal-2020-multilingual}, could help for multilingual semantic parsing.

\bibliography{aaai22,anthology}

\begin{thebibliography}{41}
\providecommand{\natexlab}[1]{#1}

\bibitem[{Aghajanyan et~al.(2020{\natexlab{a}})Aghajanyan, Maillard,
  Shrivastava, Diedrick, Haeger, Li, Mehdad, Stoyanov, Kumar, Lewis, and
  Gupta}]{aghajanyan-etal-2020-conversational}
Aghajanyan, A.; Maillard, J.; Shrivastava, A.; Diedrick, K.; Haeger, M.; Li,
  H.; Mehdad, Y.; Stoyanov, V.; Kumar, A.; Lewis, M.; and Gupta, S.
  2020{\natexlab{a}}.
\newblock Conversational Semantic Parsing.
\newblock In \emph{Proceedings of the 2020 Conference on Empirical Methods in
  Natural Language Processing (EMNLP)}, 5026--5035. Online: Association for
  Computational Linguistics.

\bibitem[{Aghajanyan et~al.(2020{\natexlab{b}})Aghajanyan, Maillard,
  Shrivastava, Diedrick, Haeger, Li, Mehdad, Stoyanov, Kumar, Lewis
  et~al.}]{aghajanyan2020conversational}
Aghajanyan, A.; Maillard, J.; Shrivastava, A.; Diedrick, K.; Haeger, M.; Li,
  H.; Mehdad, Y.; Stoyanov, V.; Kumar, A.; Lewis, M.; et~al.
  2020{\natexlab{b}}.
\newblock Conversational Semantic Parsing.
\newblock In \emph{Proceedings of the 2020 Conference on Empirical Methods in
  Natural Language Processing (EMNLP)}, 5026--5035.

\bibitem[{Bapna and Firat(2019)}]{bapna2019non}
Bapna, A.; and Firat, O. 2019.
\newblock Non-Parametric Adaptation for Neural Machine Translation.
\newblock In \emph{Proceedings of the 2019 Conference of the North American
  Chapter of the Association for Computational Linguistics: Human Language
  Technologies, Volume 1 (Long and Short Papers)}, 1921--1931.

\bibitem[{Bourtoule et~al.(2019)Bourtoule, Chandrasekaran, Choquette-Choo, Jia,
  Travers, Zhang, Lie, and Papernot}]{bourtoule2019machine}
Bourtoule, L.; Chandrasekaran, V.; Choquette-Choo, C.~A.; Jia, H.; Travers, A.;
  Zhang, B.; Lie, D.; and Papernot, N. 2019.
\newblock Machine unlearning.
\newblock \emph{arXiv preprint arXiv:1912.03817}.

\bibitem[{Chen et~al.(2018)Chen, Zhu, Ling, Inkpen, and Wei}]{chen2018neural}
Chen, Q.; Zhu, X.; Ling, Z.-H.; Inkpen, D.; and Wei, S. 2018.
\newblock Neural Natural Language Inference Models Enhanced with External
  Knowledge.
\newblock In \emph{Proceedings of the 56th Annual Meeting of the Association
  for Computational Linguistics (Volume 1: Long Papers)}, 2406--2417.

\bibitem[{Chen et~al.(2020{\natexlab{a}})Chen, Ghoshal, Mehdad, Zettlemoyer,
  and Gupta}]{chen2020low}
Chen, X.; Ghoshal, A.; Mehdad, Y.; Zettlemoyer, L.; and Gupta, S.
  2020{\natexlab{a}}.
\newblock Low-Resource Domain Adaptation for Compositional Task-Oriented
  Semantic Parsing.
\newblock In \emph{Proceedings of the 2020 Conference on Empirical Methods in
  Natural Language Processing (EMNLP)}, 5090--5100.

\bibitem[{Chen et~al.(2020{\natexlab{b}})Chen, Ghoshal, Mehdad, Zettlemoyer,
  and Gupta}]{chen-etal-2020-low}
Chen, X.; Ghoshal, A.; Mehdad, Y.; Zettlemoyer, L.; and Gupta, S.
  2020{\natexlab{b}}.
\newblock Low-Resource Domain Adaptation for Compositional Task-Oriented
  Semantic Parsing.
\newblock In \emph{Proceedings of the 2020 Conference on Empirical Methods in
  Natural Language Processing (EMNLP)}, 5090--5100. Online: Association for
  Computational Linguistics.

\bibitem[{Cheng et~al.(2020)Cheng, Agrawal, Alonso, Bhargava, Driesen, Flego,
  Kaplan, Kartsaklis, Li, Piraviperumal et~al.}]{cheng2020conversational}
Cheng, J.; Agrawal, D.; Alonso, H.~M.; Bhargava, S.; Driesen, J.; Flego, F.;
  Kaplan, D.; Kartsaklis, D.; Li, L.; Piraviperumal, D.; et~al. 2020.
\newblock Conversational Semantic Parsing for Dialog State Tracking.
\newblock In \emph{Proceedings of the 2020 Conference on Empirical Methods in
  Natural Language Processing (EMNLP)}, 8107--8117.

\bibitem[{Fan et~al.(2021)Fan, Gardent, Braud, and Bordes}]{fan2021augmenting}
Fan, A.; Gardent, C.; Braud, C.; and Bordes, A. 2021.
\newblock Augmenting Transformers with KNN-Based Composite Memory for Dialog.
\newblock \emph{Transactions of the Association for Computational Linguistics},
  9: 82--99.

\bibitem[{Ghoshal et~al.(2021)Ghoshal, Chen, Gupta, Zettlemoyer, and
  Mehdad}]{ghoshal2021learning}
Ghoshal, A.; Chen, X.; Gupta, S.; Zettlemoyer, L.; and Mehdad, Y. 2021.
\newblock Learning Better Structured Representations Using Low-rank Adaptive
  Label Smoothing.
\newblock In \emph{International Conference on Learning Representations}.

\bibitem[{Gupta et~al.(2018)Gupta, Shah, Mohit, Kumar, and
  Lewis}]{gupta2018semantic}
Gupta, S.; Shah, R.; Mohit, M.; Kumar, A.; and Lewis, M. 2018.
\newblock Semantic Parsing for Task Oriented Dialog using Hierarchical
  Representations.
\newblock In \emph{Proceedings of the 2018 Conference on Empirical Methods in
  Natural Language Processing}, 2787--2792.

\bibitem[{Guu et~al.(2020)Guu, Lee, Tung, Pasupat, and Chang}]{guu2020realm}
Guu, K.; Lee, K.; Tung, Z.; Pasupat, P.; and Chang, M.-W. 2020.
\newblock Realm: Retrieval-augmented language model pre-training.
\newblock \emph{arXiv preprint arXiv:2002.08909}.

\bibitem[{Johnson, Douze, and J{\'e}gou(2017)}]{JDH17}
Johnson, J.; Douze, M.; and J{\'e}gou, H. 2017.
\newblock Billion-scale similarity search with GPUs.
\newblock \emph{arXiv preprint arXiv:1702.08734}.

\bibitem[{Johnson, Douze, and J{\'e}gou(2019)}]{johnson2019billion}
Johnson, J.; Douze, M.; and J{\'e}gou, H. 2019.
\newblock Billion-scale similarity search with gpus.
\newblock \emph{IEEE Transactions on Big Data}.

\bibitem[{Khandelwal et~al.(2020{\natexlab{a}})Khandelwal, Fan, Jurafsky,
  Zettlemoyer, and Lewis}]{khandelwal2020nearest}
Khandelwal, U.; Fan, A.; Jurafsky, D.; Zettlemoyer, L.; and Lewis, M.
  2020{\natexlab{a}}.
\newblock Nearest neighbor machine translation.
\newblock \emph{arXiv preprint arXiv:2010.00710}.

\bibitem[{Khandelwal et~al.(2019)Khandelwal, Levy, Jurafsky, Zettlemoyer, and
  Lewis}]{khandelwal2019generalization}
Khandelwal, U.; Levy, O.; Jurafsky, D.; Zettlemoyer, L.; and Lewis, M. 2019.
\newblock Generalization through Memorization: Nearest Neighbor Language
  Models.
\newblock In \emph{International Conference on Learning Representations}.

\bibitem[{Khandelwal et~al.(2020{\natexlab{b}})Khandelwal, Levy, Jurafsky,
  Zettlemoyer, and Lewis}]{Khandelwal2020Generalization}
Khandelwal, U.; Levy, O.; Jurafsky, D.; Zettlemoyer, L.; and Lewis, M.
  2020{\natexlab{b}}.
\newblock Generalization through Memorization: Nearest Neighbor Language
  Models.
\newblock In \emph{International Conference on Learning Representations}.

\bibitem[{Kingma and Ba(2014)}]{kingma2014adam}
Kingma, D.~P.; and Ba, J. 2014.
\newblock Adam: A method for stochastic optimization.
\newblock \emph{arXiv preprint arXiv:1412.6980}.

\bibitem[{Lewis et~al.(2020{\natexlab{a}})Lewis, Ghazvininejad, Ghosh,
  Aghajanyan, Wang, and Zettlemoyer}]{lewis2020pre}
Lewis, M.; Ghazvininejad, M.; Ghosh, G.; Aghajanyan, A.; Wang, S.; and
  Zettlemoyer, L. 2020{\natexlab{a}}.
\newblock Pre-training via Paraphrasing.
\newblock \emph{Advances in Neural Information Processing Systems}, 33.

\bibitem[{Lewis et~al.(2020{\natexlab{b}})Lewis, Liu, Goyal, Ghazvininejad,
  Mohamed, Levy, Stoyanov, and Zettlemoyer}]{lewis2020bart}
Lewis, M.; Liu, Y.; Goyal, N.; Ghazvininejad, M.; Mohamed, A.; Levy, O.;
  Stoyanov, V.; and Zettlemoyer, L. 2020{\natexlab{b}}.
\newblock BART: Denoising Sequence-to-Sequence Pre-training for Natural
  Language Generation, Translation, and Comprehension.
\newblock In \emph{Proceedings of the 58th Annual Meeting of the Association
  for Computational Linguistics}, 7871--7880.

\bibitem[{Lewis et~al.(2020{\natexlab{c}})Lewis, Liu, Goyal, Ghazvininejad,
  Mohamed, Levy, Stoyanov, and Zettlemoyer}]{lewis-etal-2020-bart}
Lewis, M.; Liu, Y.; Goyal, N.; Ghazvininejad, M.; Mohamed, A.; Levy, O.;
  Stoyanov, V.; and Zettlemoyer, L. 2020{\natexlab{c}}.
\newblock {BART}: Denoising Sequence-to-Sequence Pre-training for Natural
  Language Generation, Translation, and Comprehension.
\newblock In \emph{Proceedings of the 58th Annual Meeting of the Association
  for Computational Linguistics}, 7871--7880. Online: Association for
  Computational Linguistics.

\bibitem[{Lewis et~al.(2020{\natexlab{d}})Lewis, Perez, Piktus, Petroni,
  Karpukhin, Goyal, K{\"u}ttler, Lewis, Yih, Rockt{\"a}schel
  et~al.}]{lewis2020retrieval}
Lewis, P.; Perez, E.; Piktus, A.; Petroni, F.; Karpukhin, V.; Goyal, N.;
  K{\"u}ttler, H.; Lewis, M.; Yih, W.-t.; Rockt{\"a}schel, T.; et~al.
  2020{\natexlab{d}}.
\newblock Retrieval-augmented generation for knowledge-intensive nlp tasks.
\newblock \emph{arXiv preprint arXiv:2005.11401}.

\bibitem[{Lewis et~al.(2020{\natexlab{e}})Lewis, Perez, Piktus, Petroni,
  Karpukhin, Goyal, Küttler, Lewis, tau Yih, Rocktäschel, Riedel, and
  Kiela}]{lewis2020retrievalaugmented}
Lewis, P.; Perez, E.; Piktus, A.; Petroni, F.; Karpukhin, V.; Goyal, N.;
  Küttler, H.; Lewis, M.; tau Yih, W.; Rocktäschel, T.; Riedel, S.; and
  Kiela, D. 2020{\natexlab{e}}.
\newblock Retrieval-Augmented Generation for Knowledge-Intensive NLP Tasks.
\newblock arXiv:2005.11401.

\bibitem[{Li et~al.(2020)Li, Arora, Chen, Gupta, Gupta, and
  Mehdad}]{li2020mtop}
Li, H.; Arora, A.; Chen, S.; Gupta, A.; Gupta, S.; and Mehdad, Y. 2020.
\newblock MTOP: A comprehensive multilingual task-oriented semantic parsing
  benchmark.
\newblock \emph{arXiv preprint arXiv:2008.09335}.

\bibitem[{Liu, Shi, and Chen(2020)}]{liu-etal-2020-multilingual}
Liu, Z.; Shi, K.; and Chen, N. 2020.
\newblock Multilingual Neural {RST} Discourse Parsing.
\newblock In \emph{Proceedings of the 28th International Conference on
  Computational Linguistics}, 6730--6738. Barcelona, Spain (Online):
  International Committee on Computational Linguistics.

\bibitem[{Louvan and Magnini(2020)}]{louvan2020recent}
Louvan, S.; and Magnini, B. 2020.
\newblock Recent Neural Methods on Slot Filling and Intent Classification for
  Task-Oriented Dialogue Systems: A Survey.
\newblock In \emph{Proceedings of the 28th International Conference on
  Computational Linguistics}, 480--496.

\bibitem[{Mehri, Eric, and Hakkani-Tur(2020)}]{mehri2020dialoglue}
Mehri, S.; Eric, M.; and Hakkani-Tur, D. 2020.
\newblock Dialoglue: A natural language understanding benchmark for
  task-oriented dialogue.
\newblock \emph{arXiv preprint arXiv:2009.13570}.

\bibitem[{Neeraja, Gupta, and Srikumar(2021)}]{neeraja-etal-2021-infotabskg}
Neeraja, J.; Gupta, V.; and Srikumar, V. 2021.
\newblock Incorporating External Knowledge to Enhance Tabular Reasoning.
\newblock In \emph{Proceedings of the 2021 Conference of the North American
  Chapter of the Association for Computational Linguistics: Human Language
  Technologies}. Online: Association for Computational Linguistics.

\bibitem[{Papernot and McDaniel(2018)}]{papernot2018deep}
Papernot, N.; and McDaniel, P. 2018.
\newblock Deep k-nearest neighbors: Towards confident, interpretable and robust
  deep learning.
\newblock \emph{arXiv preprint arXiv:1803.04765}.

\bibitem[{Papineni et~al.(2002)Papineni, Roukos, Ward, and
  Zhu}]{papineni2002bleu}
Papineni, K.; Roukos, S.; Ward, T.; and Zhu, W.-J. 2002.
\newblock Bleu: a method for automatic evaluation of machine translation.
\newblock In \emph{Proceedings of the 40th annual meeting of the Association
  for Computational Linguistics}, 311--318.

\bibitem[{Raffel et~al.(2020)Raffel, Shazeer, Roberts, Lee, Narang, Matena,
  Zhou, Li, and Liu}]{raffel2019exploring}
Raffel, C.; Shazeer, N.; Roberts, A.; Lee, K.; Narang, S.; Matena, M.; Zhou,
  Y.; Li, W.; and Liu, P.~J. 2020.
\newblock Exploring the Limits of Transfer Learning with a Unified Text-to-Text
  Transformer.
\newblock \emph{Journal of Machine Learning Research}, 21: 1--67.

\bibitem[{Roberts, Raffel, and Shazeer(2020)}]{roberts2020knowledge}
Roberts, A.; Raffel, C.; and Shazeer, N. 2020.
\newblock How Much Knowledge Can You Pack Into the Parameters of a Language
  Model?
\newblock arXiv:2002.08910.

\bibitem[{Rongali et~al.(2020)Rongali, Soldaini, Monti, and
  Hamza}]{rongali2020don}
Rongali, S.; Soldaini, L.; Monti, E.; and Hamza, W. 2020.
\newblock Don’t parse, generate! a sequence to sequence architecture for
  task-oriented semantic parsing.
\newblock In \emph{Proceedings of The Web Conference 2020}, 2962--2968.

\bibitem[{See, Liu, and Manning(2017)}]{see2017get}
See, A.; Liu, P.~J.; and Manning, C.~D. 2017.
\newblock Get To The Point: Summarization with Pointer-Generator Networks.
\newblock In \emph{Proceedings of the 55th Annual Meeting of the Association
  for Computational Linguistics (Volume 1: Long Papers)}, 1073--1083.

\bibitem[{Sukhbaatar et~al.(2015{\natexlab{a}})Sukhbaatar, szlam, Weston, and
  Fergus}]{sukhnaatar2015mem}
Sukhbaatar, S.; szlam, a.; Weston, J.; and Fergus, R. 2015{\natexlab{a}}.
\newblock End-To-End Memory Networks.
\newblock In Cortes, C.; Lawrence, N.; Lee, D.; Sugiyama, M.; and Garnett, R.,
  eds., \emph{Advances in Neural Information Processing Systems}, volume~28.
  Curran Associates, Inc.

\bibitem[{Sukhbaatar et~al.(2015{\natexlab{b}})Sukhbaatar, Szlam, Weston, and
  Fergus}]{sukhbaatar2015end}
Sukhbaatar, S.; Szlam, A.; Weston, J.; and Fergus, R. 2015{\natexlab{b}}.
\newblock End-to-end memory networks.
\newblock In \emph{Proceedings of the 28th International Conference on Neural
  Information Processing Systems-Volume 2}, 2440--2448.

\bibitem[{Sun et~al.(2019)Sun, Tang, Xu, Duan, Feng, Qin, Liu, and
  Zhou}]{sun2019knowledge}
Sun, Y.; Tang, D.; Xu, J.; Duan, N.; Feng, X.; Qin, B.; Liu, T.; and Zhou, M.
  2019.
\newblock Knowledge-aware conversational semantic parsing over web tables.
\newblock In \emph{CCF International Conference on Natural Language Processing
  and Chinese Computing}, 827--839. Springer.

\bibitem[{Tran et~al.(2020)Tran, Tang, Li, and Gu}]{tran2020cross}
Tran, C.; Tang, Y.; Li, X.; and Gu, J. 2020.
\newblock Cross-lingual retrieval for iterative self-supervised training.
\newblock \emph{arXiv preprint arXiv:2006.09526}.

\bibitem[{Wang et~al.(2019)Wang, Kapanipathi, Musa, Yu, Talamadupula,
  Abdelaziz, Chang, Fokoue, Makni, Mattei et~al.}]{wang2019improving}
Wang, X.; Kapanipathi, P.; Musa, R.; Yu, M.; Talamadupula, K.; Abdelaziz, I.;
  Chang, M.; Fokoue, A.; Makni, B.; Mattei, N.; et~al. 2019.
\newblock Improving natural language inference using external knowledge in the
  science questions domain.
\newblock In \emph{Proceedings of the AAAI Conference on Artificial
  Intelligence}, volume 33/01, 7208--7215.

\bibitem[{Weston, Copra, and Bordes(2014)}]{weston2014memory}
Weston, J.; Copra, S.; and Bordes, A. 2014.
\newblock Memory networks.
\newblock \emph{arXiv preprint arXiv:1410.3916}.

\bibitem[{Weston, Dinan, and Miller(2018)}]{weston2018retrieve}
Weston, J.; Dinan, E.; and Miller, A. 2018.
\newblock Retrieve and Refine: Improved Sequence Generation Models For
  Dialogue.
\newblock In \emph{Proceedings of the 2018 EMNLP Workshop SCAI: The 2nd
  International Workshop on Search-Oriented Conversational AI}, 87--92.

\end{thebibliography}

\clearpage
\section{Qualitative Examples}
\label{sec:qualitative_examples}
Below, we provide some qualiative example from the TOPv2 datasets, in the first box, we shows the input utterance, nearest neighbour utterance i.e. nearest-nn utterance and it corresponding semantic parse i.e. semparse-nn (from the training set). In the second box we showed expected gold semantic parse (expected-sparse) and the semantic parse prediction with baseline i.e. without nearest neighbor (without-nn), and our models utterance only nearest neighbor, where the input is appending with nearest-nn utterance (utterance-nn) and semantic augmented model i.e. (semparse-nn), where the input is augmented with semantic parse of first nearest neighbor. We can clearly see from the examples that augmented nearest neighbor enhance models in identifying correct intent and slots.

\noindent \exampleParagraph{example 1 input\\}
{

\noindent \textbf{utterance : }{message just lizzie and trent from my group if they have any updates yet ?} \\

\noindent \textbf{nearest-nn utterance : }{message kira and lena saying want to get drinks this week ?} 

\noindent \textbf{nearest-nn semparse : }{[in:send$\_$message message [sl:recipient kira ] and [sl:recipient lena ] saying [sl:content$\_$exact want to get drinks this week ]?]}\\
}

\noindent \exampleParagraph{example 1 output\\}
{

\noindent  \textbf{expected-parse : }{[in:send$\_$message [sl:recipient lizzie ] [sl:recipient trent ] [sl:content$\_$exact they have any updates yet ] ]}\\

\noindent \textbf{without-nn : }{[\textcolor{red}{in:get$\_$message} [sl:content$\_$exact they have any updates yet ] [\textcolor{red}{sl:group} lizzie ] \textcolor{red}{[sl:group} trent ] ]} \\

\noindent \textbf{semparse-nn : }{[in:send$\_$message [sl:recipient lizzie ] [sl:recipient trent ] [sl:content$\_$exact they have any updates yet ] ]}
\\

\noindent \textbf{utterance-nn : }{[in:send$\_$message [sl:recipient lizzie ] [sl:recipient trent ] [sl:content$\_$exact they have any updates yet ] ]}\\

}

\noindent In example $1$, the model misses the correct intent and corresponding slots completely, the correct intent is sending a message rather than receiving a message is correctly identified by both semparse-nn and utterance-nn.\\

\noindent \exampleParagraph{example 2 input\\}
{

\noindent \textbf{utterance : }{no more country} \\

\noindent \textbf{nearest-nn utterance : }{no more music} \\

\noindent \textbf{nearest-nn semparse : }{[in:stop$\_$music [sl:music$\_$type music ] ]}\\

}

\noindent \exampleParagraph{example 2 output\\}
{

\noindent  \textbf{expected-parse : }{[in:remove$\_$from$\_$playlist$\_$music [sl:music$\_$genre country ] ]}\\

\noindent \textbf{without-nn : }{[\textcolor{red}{in:play$\_$music} [sl:music$\_$genre country ] ]} \\

\noindent \textbf{semparse-nn : }{[in:remove$\_$from$\_$playlist$\_$music [sl:music$\_$genre country ] ]}
\\

\noindent \textbf{utterance-nn : }{[in:remove$\_$from$\_$playlist$\_$music [sl:music$\_$genre country ] ]}\\

}\\

\noindent In example $2$, the baseline model without nearest neighbour did the exact opposite of intended task of removing music of genre country from the playlist. However, after augmenting nearest neighbor context the model quickly correct the indentd intent and slot. It should also be noted the both the correct intent and slot (i.e. in:remove$\_$from$\_$playlist$\_$music and sl:music$\_$genre) are not present in the nearest-nn semparse  but it do contain similar intent and slot (i.e. in:stop$\_$music. and sl:music$\_$type), which help retrieval augmented model in correct prediction. As earlier the model is able to predict correct even with utterance only augmentation too.

\noindent \exampleParagraph{example 3 input\\}
{

\noindent \textbf{utterance : }{block all songs of mariah carey} \\

\noindent \textbf{nearest-nn utterance : }{delete mariah carey songs} \\

\noindent \textbf{nearest-nn semparse : }{[in:remove$\_$from$\_$playlist$\_$music delete [sl:music$\_$artist$\_$name mariah carey] [sl:music$\_$type songs ] ]}\\
}\\\\

\noindent \exampleParagraph{example 3 output\\}
{

\noindent  \textbf{expected-parse : }{[in:remove$\_$from$\_$playlist$\_$music [sl:music$\_$artist$\_$name mariah carey ] ]}\\

\noindent \textbf{without-nn : }{[\textcolor{red}{in:unsupported$\_$music} [\textcolor{red}{sl:music$\_$type} songs ] ]} \\

\noindent \textbf{semparse-nn : }{[in:remove$\_$from$\_$playlist$\_$music [sl:music$\_$type songs ] [sl:music$\_$artist$\_$name mariah carey ] ]}\\\\

\noindent \textbf{utterance-nn : }{[in:remove$\_$from$\_$playlist$\_$music [sl:music$\_$type songs ] [sl:music$\_$artist$\_$name mariah carey ] ]}\\
} \\\\

\noindent In example 3 the model without nearest neighbor augmentation struggle to identify the intent from utterance text token ``block" therefore prediction unsupported music as the intent and the music type as songs, however the model with augmented nearest neighbour example with ``delete" intended slot correct identified both the intent and slots. Furthermore, the model also resolve the active passive voice confusion with nearest neighbor augmentation.

\section{Domain based Limited Training Setting}
\label{sec:domainlimitsetting}

In Table \ref{tab:limiteddomain} shows the performance of model for each domain on original baseline (without-nn), and RetroNLU model utterance-nn and semparse-nn with varying amount of supervised training data. Overall, semparse-nn outperform utterance-nn over most of the domains. Surprising, we also found that for few domain (with large number of samples) utterance-nn perform marginally better than semparse-nn, need to investigate exact reason for that. As expected both model utternace-nn and semparse-nn perform much better than original baseline which is without any nearest neighbour augmentation.

\begin{table*}[!htbp]
    \centering
    \begin{tabular}{c|ccc|ccc|ccc}
    \toprule
    \multicolumn{1}{c}{\bf Percentage} & \multicolumn{3}{c}{\bf 10 $\%$} & \multicolumn{3}{c}{\bf 20 $\%$} & \multicolumn{3}{c}{\bf 30 $\%$} \\ \hdashline
\bf Domain &\bf w/o nn &\bf uttr-nn &\bf sem-nn &\bf w/o nn &\bf uttr-nn &\bf sem-nn &\bf w/o nn &\bf uttr-nn &\bf sem-nn \\ 
\midrule
\bf Alarm & 80.50 & 84.05 & 83.60 & 83.71 & 84.89 & 85.76 & 84.22 & 85.93 & 82.92 \\
\bf Event & 68.56 & 78.33 & 79.38 & 75.01 & 80.85 & 82.32 & 77.64 & 81.91 & 82.92\\

\bf Music & 69.12 & 75.74 & 73.23 & 74.09 & 77.53 & 77.34 & 75.6 & 78.01 & 78.13 \\
Timer & 71.63 & 76.76 & 76.27 & 75.51 & 76.18 & 79.28 & 77.21 & 79.68 & 79.84 \\
\bf Navigation & 74.30 & 73.86 & 76.44 & 77.89 & 79.40 & 79.96 & 80.11 & 81.79 & 81.61 \\
\bf Messaging & 84.38 & 87.30 & 89.44 & 88.39 & 91.31 & 91.50 & 89.53 & 92.78 & 92.25 \\
\bottomrule
    \end{tabular}
    \caption{Limited training setting results on various domain with baseline (without-nn), \approach~ model utterance-nn and semparse-nn, shown here as w/o nn, utter-nn and sem-nn respectively.}
    \label{tab:limiteddomain}
\end{table*}

\section{Domain Specific Effect of Nearest Neighbours}
\label{sec:domain_neighbour}

In Table \ref{tab:numneighdomain} we shows the performance of model for each domain on original baseline (without-nn), and RetroNLU model utterance-nn and semparse-nn with varying number of nearest neighbour augmented. We found the utternace-nn performance increases with increasing number of neighbours where semparse performance remain mostly constant after the first neighbour augmentation for many domains. We suspect this is due to the fact that the data contains a large number of utterances with identical semparse output..  There is also frame redundancy, since many unique utterance inquiries have comparable semantic parse frames structure with differences only on slot values.

\begin{table*}[!htbp]
    \centering
    \begin{tabular}{c|ccc|ccc|ccc}
    \toprule
    \multicolumn{1}{c}{\bf \#neighbour's} & \multicolumn{3}{c}{\bf one} & \multicolumn{3}{c}{\bf two} & \multicolumn{3}{c}{\bf three} \\ \hdashline
\bf Domain &\bf w/o nn &\bf uttr-nn &\bf sem-nn &\bf w/o nn &\bf uttr-nn &\bf sem-nn  &\bf w/o nn &\bf uttr-nn &\bf sem-nn \\ 
\midrule
\bf Alarm & 86.67 & 87.17 & 88.57 & 86.67 & 87.77 & 87.87 & 86.67 & 87.68 & 87.90 \\
\bf Event & 83.83 & 85.03 & 84.77 & 83.83 & 84.92 & 85.26 & 83.83& 85.26 & 85.34 \\
\bf Music & 79.80 & 80.73 & 80.71 & 79.80 & 80.71 & 81.50 & 79.80 & 80.52 & 81.11 \\
\bf Timer & 81.21 & 81.75 & 81.01 & 81.21& 81.04 & 82.29 & 81.21 & 81.44 & 82.10 \\
\bf Messaging & 93.50 & 94.52 & 94.65 & 93.50 & 94.92 & 95.05 & 93.50 & 94.88 & 94.92 \\
\bf Navigation & 82.96 & 84.16 & 85.20 & 82.96 & 84.12 & 84.46 & 82.96 & 84.59 & 84.79 \\
\bottomrule
\end{tabular}
\caption{Effect of number of nearest neighbours of \approach~ performance with varying domains}
\label{tab:numneighdomain}
\end{table*}

\end{document}